\documentclass[10pt,twocolumn,letterpaper]{article}

\usepackage{cvpr}
\usepackage{times}
\usepackage{epsfig}
\usepackage{graphicx}
\usepackage{amsmath}
\usepackage{amssymb}
\usepackage{tabularx}
\usepackage{booktabs}
\usepackage[dvipsnames]{xcolor}

\usepackage[pagebackref=true,breaklinks=true,letterpaper=true,colorlinks,bookmarks=false]{hyperref}

\cvprfinalcopy 
\usepackage{multirow}
\newcommand{\mathbbm}[1]{\text{\usefont{U}{bbm}{m}{n}#1}}

\newcommand{\ma}[1]{\mathrm{#1}} 

\def\cvprPaperID{1328} 

\ifcvprfinal\fi
\begin{document}

\title{Neural Architecture Search for Lightweight Non-Local Networks}
\author{
Yingwei Li\textsuperscript{\rm 1$\ast$}~~~~
Xiaojie Jin\textsuperscript{\rm 2}~~~~
Jieru Mei\textsuperscript{\rm 1$\ast$}~~~~
Xiaochen Lian\textsuperscript{\rm 2}~~~~
Linjie Yang\textsuperscript{\rm 2}~~~~\\
Cihang Xie\textsuperscript{\rm 1}~~~~
Qihang Yu\textsuperscript{\rm 1}~~~~
Yuyin Zhou\textsuperscript{\rm 1}~~~~
Song Bai\textsuperscript{\rm 3}~~~~ 
Alan Yuille\textsuperscript{\rm 1} \\
\textsuperscript{\rm 1}Johns Hopkins University \qquad\qquad
\textsuperscript{\rm 2}ByteDance AI Lab \qquad\qquad
\textsuperscript{\rm 3}University of Oxford \\
\vspace{-.5em}
}

\maketitle
\thispagestyle{empty}

\begin{abstract}
Non-Local (NL) blocks have been widely studied in various vision tasks. However, it has been rarely explored to embed the NL blocks in mobile neural networks, mainly due to the following challenges: 1) NL blocks generally have heavy computation cost which makes it difficult to be applied in applications where computational resources are limited, and 2) it is an open problem to discover an optimal configuration to embed NL blocks into mobile
neural networks.
We propose \textbf{AutoNL} to overcome the above two obstacles. Firstly, we propose a Lightweight Non-Local (LightNL) block by squeezing the transformation operations and incorporating compact features. With the novel design choices, the proposed LightNL block is \textbf{400$\times$ computationally cheaper} than its conventional counterpart without sacrificing the performance. Secondly, by relaxing the structure of the LightNL block to be differentiable during training, we propose an efficient neural architecture search algorithm to learn an optimal configuration of LightNL blocks in an end-to-end manner. Notably, using only $32$ GPU hours, the searched AutoNL model achieves 77.7\% top-1 accuracy on ImageNet under a typical mobile setting (350M FLOPs), significantly outperforming previous mobile models including MobileNetV2 (+5.7\%), FBNet (+2.8\%) and MnasNet (+2.1\%). Code and models are available at \url{https://github.com/LiYingwei/AutoNL}.
\end{abstract}

\section{Introduction} \label{sec:intro}
Non-Local (NL) block~\cite{bello2019attention, wang2018non} aims to capture long-range dependencies in deep neural networks, which have been used in a variety of vision tasks such as video classification~\cite{wang2018non}, object detection~\cite{wang2018non}, semantic segmentation~\cite{zhao2018psanet,zhou2019multi}, image classification~\cite{bello2019attention}, and adversarial robustness~\cite{xie2019feature}. Despite the remarkable progress, the general utilization of non-local modules under resource-constrained scenarios such as mobile devices remains underexplored. This may be due to the following two factors. \let\thefootnote\relax\footnote{$^\ast$Work done during an internship at Bytedance AI Lab.}

First, NL blocks compute the response at each position by attending to all other positions
and computing a weighted average of the features in all positions, which incurs a large computation burden. Several efforts have been explored to reduce the computation overhead. For instance, \cite{chen20182,levi2018efficient} use associative law to reduce the memory and computation cost of matrix multiplication; Yue~\etal~\cite{yue2018compact} use Taylor expansion to optimize the non-local module; Cao~\etal~\cite{cao2019gcnet} compute the affinity matrix via a convolutional layer; Bello~\etal~\cite{bello2019attention} design a novel attention-augmented convolution. 
However, these methods either still lead to relatively large computation overhead (via using heavy operators, such as large matrix multiplications) or result in a less accurate outcome (\eg,~simplified NL blocks~\cite{cao2019gcnet}), making these methods undesirable for mobile-level vision systems. 

\begin{figure}[tb]
\centering
\includegraphics[width=0.85\linewidth]{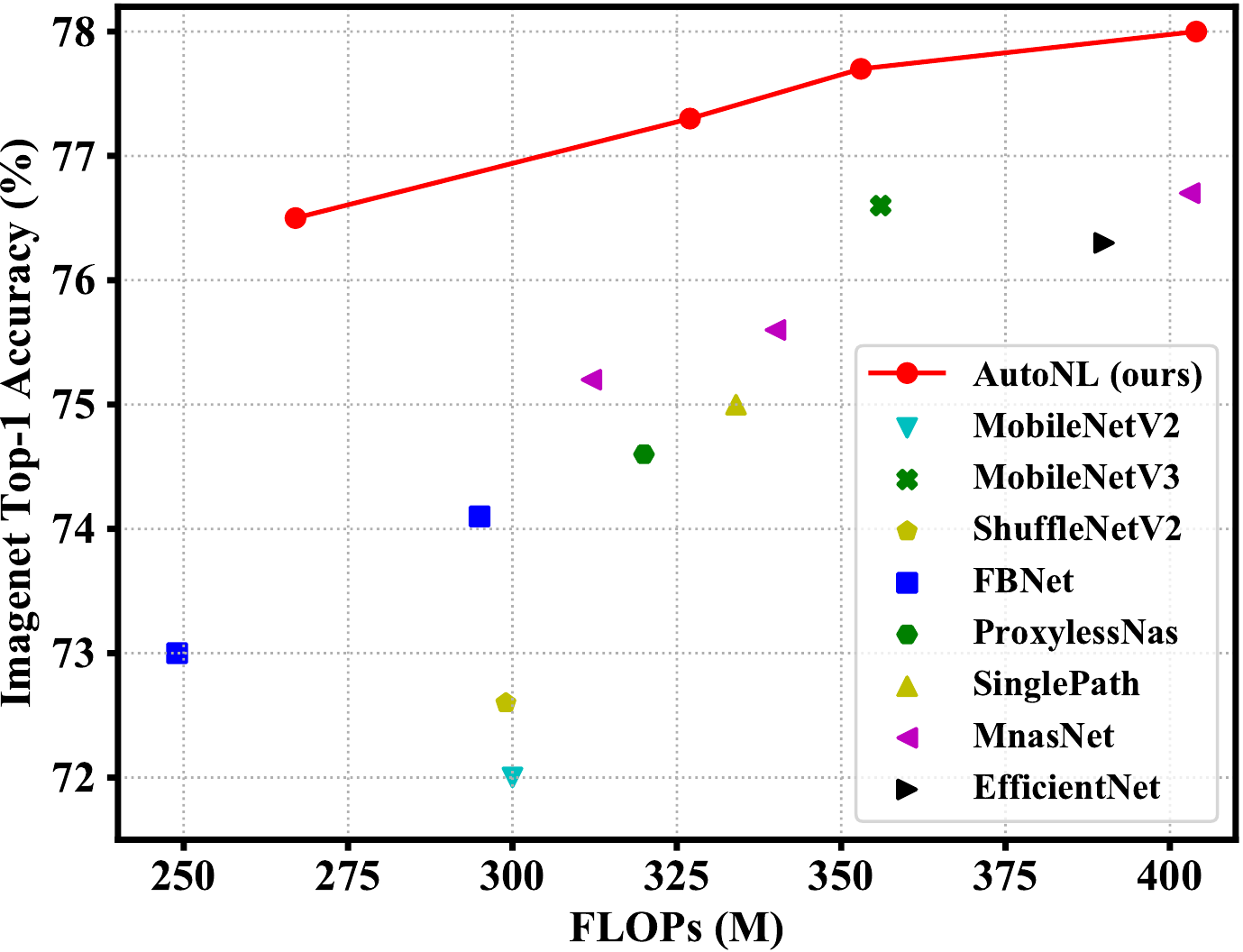}
\caption{\textbf{ImageNet Accuracy~\emph{vs.}~Computation Cost.} Details can be found in Table~\ref{tab:overall_compare}.}
\label{fig:main_acc}
\vspace{-1em}
\end{figure}

Second, NL blocks are usually implemented as individual modules which can be plugged into a few manually selected layers (usually relatively deep layers). While it is intractable to densely embed it into a deep network due to the high computational complexity, it remains unclear where to insert those modules economically. Existing methods have not fully exploited the capacity of NL blocks in relational modeling under mobile settings.

Taking the two factors aforementioned into account, we aim to answer the following questions in this work:~\emph{is it possible to develop an efficient NL block for mobile networks? What is the optimal configuration to embed those modules into mobile neural networks?} We propose \textbf{AutoNL} to address these two questions. First, we design a Lightweight Non-Local (LightNL) block, which is the first work to apply non-local techniques to mobile networks to our best knowledge. We achieve this with two critical design choices 1) lighten the transformation operators (\eg, $1\times 1$ convolutions) and 2) utilize compact features. As a result, the proposed LightNL blocks are usually \textbf{400$\times$ computationally cheaper} than conventional NL blocks~\cite{wang2018non}, which is favorable to be applied to mobile deep learning systems.
Second, we propose a novel neural architecture search algorithm. Specifically, we relax the structure of LightNL blocks to be differentiable so that our search algorithm can simultaneously determine the compactness of the features and the locations for LightNL blocks during the end-to-end training. We also reuse intermediate search results by acquiring various affinity matrices in one shot to reduce the redundant computation cost, which speeds up the search process.

Our proposed searching algorithm is fast and delivers high-performance lightweight models. As shown in Figure~\ref{fig:main_acc}, our searched small AutoNL model achieves $76.5\%$ ImageNet top-1 accuracy with $267$M FLOPs, which is faster than MobileNetV3~\cite{howard2019searching} with comparable performance ($76.6\%$ top-1 accuracy with $356$M FLOPs). Also, our searched large AutoNL model achieves $77.7\%$ ImageNet top-1 accuracy with $353$M FLOPs, which has similar computation cost as MobileNetV3 but improves the top-1 accuracy by $1.1\%$.

To summarize, our contributions are three-fold: (1) We design a lightweight and search compatible NL block for visual recognition models on mobile devices and resource-constrained platforms; (2) We propose an efficient neural architecture search algorithm to automatically learn an optimal configuration of the proposed LightNL blocks; 3) Our model achieves state-of-the-art performance on the ImageNet classification task under mobile settings.

\begin{figure*}[tb]
\centering
\includegraphics[width=0.80\linewidth]{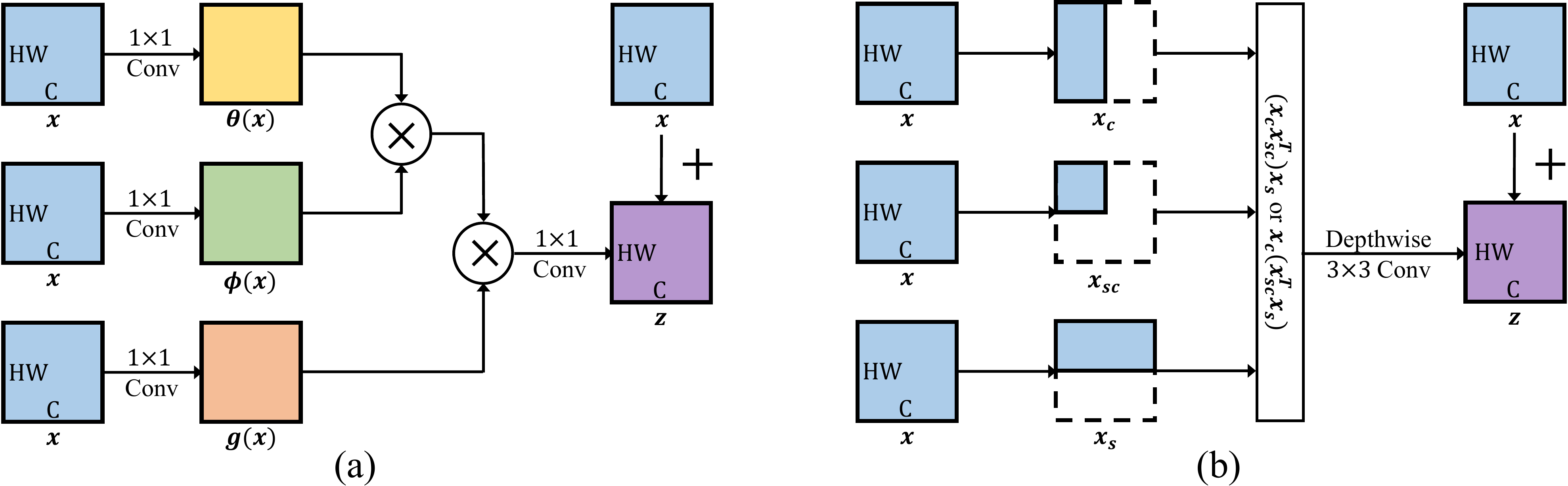}
\caption{\textbf{Original NL~\emph{vs.}~LightNL Block.} (a) A typical architecture of the NL block contains several heavy operators, such as $1\times 1$ convolution ops and large matrix multiplications. (b) The proposed LightNL block contains much more lightweight operators, such as depthwise convolution ops and small matrix multiplications.}
\label{fig:nl_lsam}
\vspace{-1em}
\end{figure*}

\section{Related Work}
\noindent\textbf{Attention mechanism.}
The attention mechanism has been successfully applied to neural language processing in recent years~\cite{bahadnau2015joint_align, vaswani2017attention, devlin2019bert}. 
Wang~\etal ~\cite{wang2018non} bridge attention mechanism and non-local operator, and use it to model long-range relationships in computer vision applications.
Attention mechanisms can be applied along two orthogonal directions: channel attention and spatial attention. Channel attention~\cite{hu2018se,wang2017residual_attention,park2018bottleneck_attention} aims to model the relationships between different channels with different semantic concepts. By focusing on a part of the channels of the input feature and deactivating non-related concepts, the models can focus on the concepts of interest. Due to its simplicity and effectiveness~\cite{hu2018se}, it is widely used in neural architecture search~\cite{tan2019mnasnet,tan2019mixnet,howard2019searching,chu2019scarletnas}.

Our work explores in both directions of spatial/channel attention. Although existing works~\cite{chen20182,yue2018compact,cao2019gcnet,bello2019attention,wang2020axial} exploit various techniques to improve efficiency, they are still too computationally heavy under mobile settings. To alleviate this problem, we design a lightweight spatial attention module with low computational cost and it can be easily integrated into mobile neural networks.

\vspace{0.5ex}\noindent\textbf{Efficient mobile architectures.}
There are a lot of handcrafted neural network architectures~\cite{huang2017condensenet,IGCV2,howard2017mobilenets,sandler2018mobilenetv2,zhang2018shufflenet,ma2018shufflenet} for mobile applications. Among them, the family of MobileNet~\cite{howard2017mobilenets,sandler2018mobilenetv2} and the family of ShuffleNet~\cite{zhang2018shufflenet,ma2018shufflenet} stand out due to their superior efficiency and performance. 
MobileNetV2~\cite{sandler2018mobilenetv2} proposes the inverted residual block to improve both efficiency and performance over MobileNetV1~\cite{howard2017mobilenets}. ShuffleNet~\cite{ma2018shufflenet} proposes to use efficient shuffle operations along with group convolutions to  design efficient networks. Above methods are usually subject to trial-and-errors by experts in the model design process.

\vspace{0.5ex}\noindent\textbf{Neural Architecture Search.}
Recently, it has received much attention to use neural architecture search (NAS) to design efficient network architectures for various applications~\cite{tan2019efficientnet,ghiasi2019fpn,liu2019auto,yu2019c2fnas,jin2019adabits}.
A critical part of NAS is to design proper search spaces.
Guided by a meta-controller, early NAS methods either use reinforcement learning~\cite{zoph2017nasnet} or evolution algorithm~\cite{real2019amoebanet} to discover better architectures. These methods are computationally inefficient, requiring thousands of GPU days to search. ENAS~\cite{hieu2018enas} shares parameters across sampled architectures to reduce the search cost. DARTS~\cite{liu2018darts} proposes a continuous relaxation of the architecture parameters and conducts one-shot search and evaluation. These methods all adopt a NASNet~\cite{zoph2017nasnet} like search space. Recently, more expert knowledge in handcrafting network architectures are introduced in NAS. Using MobileNetV2 basic blocks in search space~\cite{han2019proxyless,wu2019fbnet,tan2019mnasnet,tan2019mixnet,stamoulis2019single,zichao2019uniform_sampling,mei2020atomnas} significantly improves the performance of searched architectures. \cite{han2019proxyless,zichao2019uniform_sampling} reduce the GPU memory consumption by executing only part of the super-net in each forward pass during training.
\cite{mei2020atomnas} proposes an ensemble perspective of the basic block and simultaneously searches and trains the target architecture in the fine-grained search space.
\cite{stamoulis2019single} proposes a super-kernel representation to incorporate all architectural hyper-parameters (\eg,~kernel sizes, expansion rations in MobileNetV2 blocks) in a unified search framework to reuse model parameters and computations. 
In our proposed searching algorithm, we focus on seeking an optimal configuration of LightNL blocks in low-cost neural networks which brought significant performance gains. 


\section{AutoNL}
In this section, we present AutoNL: we first elaborate on how to design a Lightweight Non-Local (LightNL) block in Section~\ref{sec:LSAM}; 
then we introduce a novel neural architecture search algorithm in Section~\ref{sec:search} to automatically search for an optimal configuration of LightNL blocks.

\subsection{Lightweight Non-Local Blocks}
\label{sec:LSAM}
In this section, we first revisit the NL blocks, then we introduce our proposed Lightweight Non-Local (LightNL) block in detail.

\vspace{0.5ex}\noindent\textbf{Revisit NL blocks.} The core component in the NL blocks is the non-local operation. Following~\cite{wang2018non}, a generic non-local operation can be formulated as
\begin{equation} \label{eq:ori_nl}
    \ma{y}_i = \frac{1}{\mathcal{C(\ma{x})}} \sum_{\forall j}f(\ma{x}_i, \ma{x}_j)g(\ma{x}_j),
\end{equation}
where $i$ indexes the position of input feature $\ma{x}$ whose response is to be computed, $j$ enumerates all possible positions in $\ma{x}$, $f(\ma{x}_i, \ma{x}_j)$ outputs the affinity matrix between $\ma{x}_i$ and its context features $\ma{x}_j$, $g(\ma{x}_j)$ computes an embedding of the input feature at the position $j$, and $C(\ma{x})$ is the normalization term. Following~\cite{wang2018non}, the non-local operation in Eqn.~\eqref{eq:ori_nl} is wrapped into a NL block with a residual connection from the input feature $\ma{x}$. The mathematical formulation is given as
\begin{equation} \label{eq:residual}
    \ma{z}_i = W_z\ma{y}_i + \ma{x}_i,
\end{equation}
where $W_z$ denotes a learnable feature transformation.

\vspace{0.5ex}\noindent\textbf{Instantiation.}
Dot product is used as the function form of $f(x_i, x_j)$ due to its simplicity in computing the correlation between features. Eqn.~\eqref{eq:ori_nl} thus becomes
\begin{equation} \label{eq:matrix_nl}
    \ma{y}=\frac{1}{\mathcal{C}(\ma{x})}\theta(\ma{x})\theta(\ma{x})^\mathrm{T}g(\ma{x}).
\end{equation}
Here the shape of $\ma{x}$ is denoted as $(H, W, C)$ where $H$, $W$ and $C$ are the height, width and number of channels, respectively. $\theta(\cdot)$ and $g(\cdot)$ are $1\times 1$ convolutional layers with $C$ filters. Before matrix multiplications, the outputs of $1\times 1$ convolution are reshaped to $(H\times W, C)$. 

Levi~\etal~\cite{levi2018efficient} discover that for NL blocks instantiated in the form of Eqn.~\eqref{eq:matrix_nl}, employing the associative law of matrix multiplication can largely reduce the computation overhead. Based on the associative rules, Eqn.~\eqref{eq:matrix_nl} can be written in two equivalent forms:
\begin{equation} \label{eq:associative_law}\small
    \ma{y}=\frac{1}{\mathcal{C}(\ma{x})} \left(\theta(\ma{x})\theta(\ma{x})^\mathrm{T}\right)g(\ma{x}) = \frac{1}{\mathcal{C}(\ma{x})} \theta(\ma{x})\left(\theta(\ma{x})^\mathrm{T}g(\ma{x})\right).
\end{equation}
Although the two forms produce the same numerical results, they have different computational complexity~\cite{levi2018efficient}. Therefore in computing Eqn.~\eqref{eq:matrix_nl}, one can always choose the form with smaller computation cost for better efficiency. 

\vspace{0.5ex}\noindent\textbf{Design principles.}
The following part introduces two key principles to reduce the computation cost of Eqn.~\eqref{eq:matrix_nl}.

\vspace{0.5ex}\noindent\textit{Design principle 1: Share and lighten the feature transformations.}
Instead of using two different transformations ($\theta$ and $g$) on the same input feature x in Eqn.~\eqref{eq:matrix_nl}, we use a shared transformation in the non-local operation. In this way, the computation cost of Eqn.~\eqref{eq:matrix_nl} is significantly reduced by reusing the result of $g(\ma{x})$ in computing the affinity matrix. The simplified non-local operation is 
\begin{equation} \label{eq:share_nl}
    \ma{y}=\frac{1}{\mathcal{C}(\ma{x})}g(\ma{x})g(\ma{x})^\mathrm{T}g(\ma{x}).
\end{equation}

The input feature $\ma{x}$ (output of hidden layer) can be seen as the transformation of input data $\ma{x_0}$ through a feature transformer $F(\cdot)$. Therefore Eqn.~\eqref{eq:share_nl} can be written as
\begin{equation}\label{eq:simple_g_F}
    \ma{y}=\frac{1}{\mathcal{C}(F(\ma{x_0}))}g(F(\ma{x_0}))g(F(\ma{x_0}))^\mathrm{T}g(F(\ma{x_0})).
\end{equation}
In the scenario of using NL blocks in neural networks, $F(\cdot)$ is represented by a parameterized deep neural network. In contrast, $g(\cdot)$ is a single convolution operation. To further simplify Eqn.~\eqref{eq:simple_g_F}, we integrate the learning process of $g(\cdot)$ into that of $F(\cdot)$. Taking advantage of the strong capability of deep neural networks on approximating functions~\cite{hornik1989multilayer}, we remove $g(\cdot)$ and Eqn.~\eqref{eq:simple_g_F} is simplified as
\begin{equation} \label{eq:remove_nl}
    \ma{y}=\frac{1}{\mathcal{C}(\ma{x})}\ma{x}\ma{x}^\mathrm{T}\ma{x}.
\end{equation}

At last, we introduce our method to simplify ``$W_z$", another heavy
transformation function in Eqn.~\eqref{eq:residual}. 
Recent works~\cite{wang2018non} instantiate it as a $1\times1$ convolutional layer. To further reduce the computation cost of NL blocks, we propose to replace the $1\times1$ convolution with a $3\times 3$ depthwise convolution~\cite{howard2017mobilenets} since the latter is more efficient. Eqn.~\eqref{eq:residual} is then modified to be
\begin{equation} \label{eq:depthwise}
    \ma{z} = \operatorname{DepthwiseConv}(\ma{y}, W_d) + \ma{x},
\end{equation}
where $W_d$ denotes the depthwise convolution kernel. 

\vspace{0.5ex}\noindent\textit{Design principle 2: Use compact features for computing affinity matrices.}
Since $\ma{x}$ is a high-dimensional feature, directly performing matrix multiplication using the full-sized $\ma{x}$ per Eqn.~\eqref{eq:remove_nl} leads to large computation overhead. To solve this problem, we propose to downsample $\ma{x}$ first to obtain a more compact feature which replaces $\ma{x}$ in Eqn.~\eqref{eq:remove_nl}. Since $\ma{x}$ is a three-dimensional feature with depth  (channels), width and height, 
we propose to downsample $\ma{x}$ along either \textbf{channel dimension}, \textbf{spatial dimension} or \textbf{both dimensions} to obtain compact features $\ma{x_c}$, $\ma{x_s}$ and $\ma{x_{sc}}$ respectively. Consequently, the computation cost of Eqn.~\eqref{eq:remove_nl} is reduced.

Therefore, based on Eqn.~\eqref{eq:remove_nl}, we can simply apply the compact features $\{\ma{x_c}, \ma{x_{sc}}, \ma{x_s}\}$ in the NL block to compute $\ma{x_c}\ma{x_{sc}^T}$ and $\ma{x_{sc}^T}\ma{x_s}$ as
\begin{equation} \label{eq:compact_nl}
    y=\frac{1}{\mathcal{C}(\ma{x})}\ma{x}_c \ma{x}_{sc}^\mathrm{T}\ma{x}_s.
\end{equation}
Note that there is a trade-off between the computation cost and the representation capacity of the output (\ie,~$\ma{y}$) of the non-local operation: using more compact features (with a lower downsampling ratio) reduces the computation cost but the output fails to capture the informative context information in those discarded features; on the other hand, using denser features (with a higher downsampling ratio) helps the output capture richer contexts, but it is more computationally demanding. Manually setting the downsampling ratios requires trial-and-errors. To solve this issue, we propose a novel neural network architecture search (NAS) method in Section~\ref{sec:search} to efficiently search for the configuration of NL blocks that achieve descent performance under specific resource constraints. 

Before introduce our NAS method, let's briefly summarize the advantages of the proposed LightNL blocks. Thanks to the aforementioned two design principles, our proposed LightNL block is empirically demonstrated to be much  more  efficient (refer to Section~\ref{sec:exp}) than the conventional NL block~\cite{wang2018non}, making it favorable to be deployed in mobile devices with limited computational budgets. In addition, since the computational complexity of the blocks can be easily adjusted by the downsampling ratios, the proposed LightNL blocks can provide better support on deep learning models at different scales. We illustrate the structure of the conventional NL block and that of the proposed block in Figure~\ref{fig:nl_lsam}.

\begin{figure}[tb]
\centering
\includegraphics[width=0.80\linewidth]{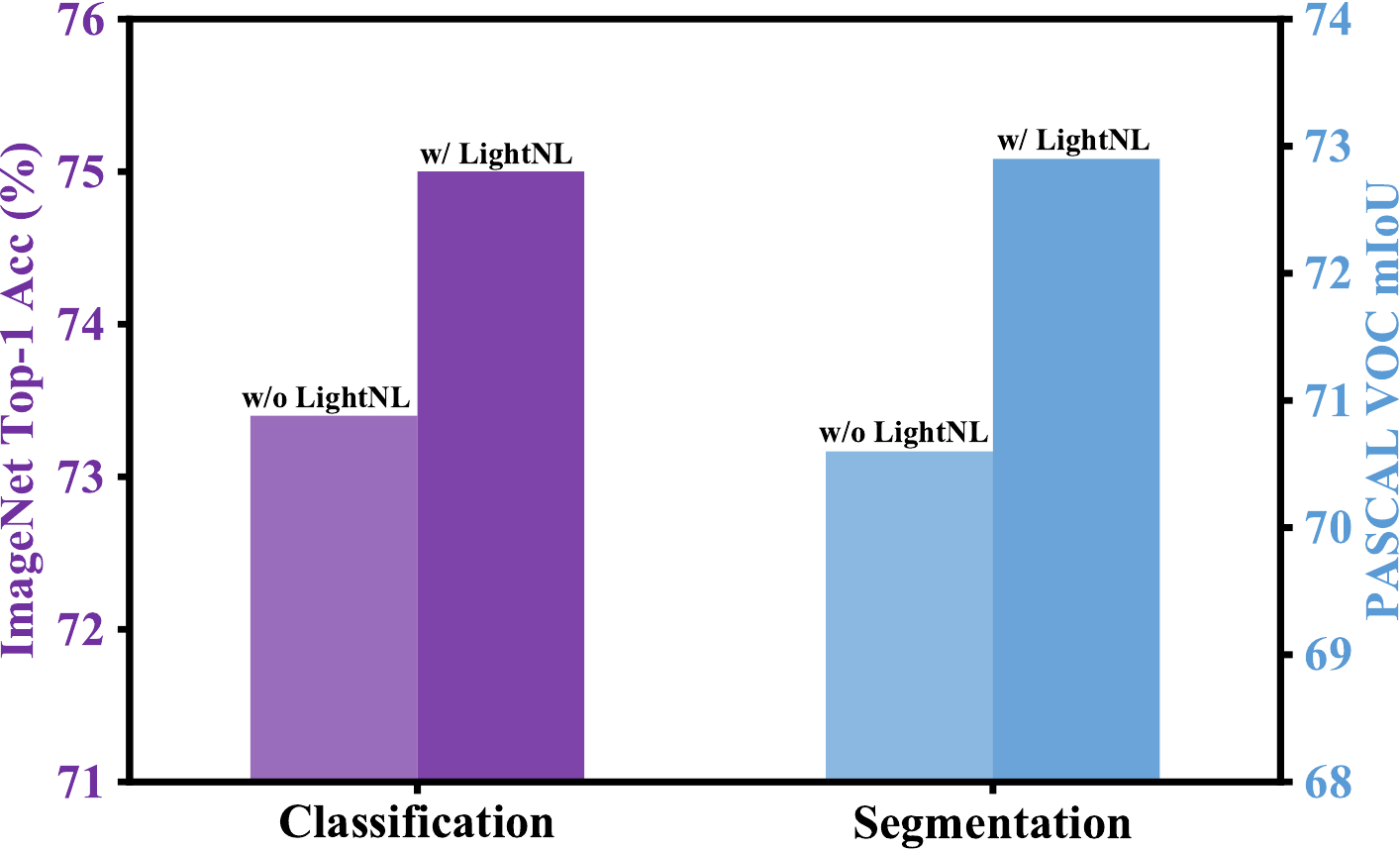}
\caption{\textbf{MobileNetV2 \emph{vs.} MobileNetV2 + LightNL.} The proposed LightNL block improves the baseline by $1.6\%$ in ImageNet top-1 accuracy and $2.3\%$ in PASCAL VOC 2012 mIoU.}
\label{fig:clsseg}
\vspace{-1em}
\end{figure}

\begin{figure*}[tb]
\centering
\includegraphics[width=0.98\linewidth]{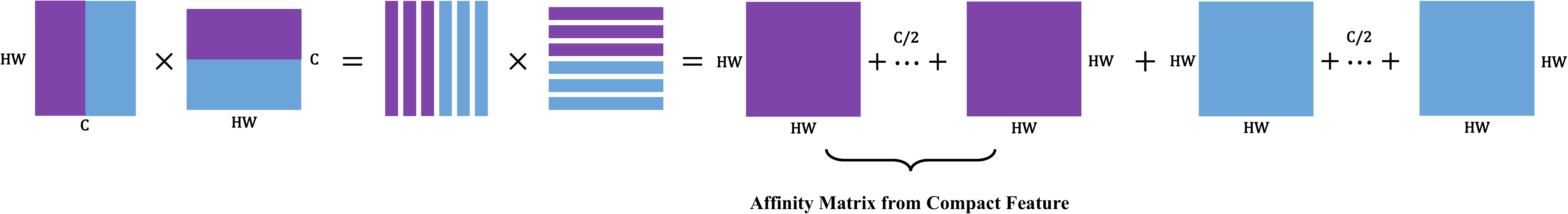}
\caption{Illustrate the feature reuse paradigm along channel dimension.}
\label{fig:fig3}
\vspace{-1em}
\end{figure*}

\subsection{Neural Architecture Search for LightNL}
\label{sec:search}

To validate the efficacy and generalization of the proposed LightNL block for deep networks, we perform a proof test by applying it to every MobileNetV2 block. As shown in Figure~\ref{fig:clsseg}, such a simple way of using the proposed LightNL blocks can already significantly boost the performance on both image classification and semantic segmentation. This observation motivates us to search for a better configuration of the proposed LightNL blocks in neural networks to fully utilize its representation learning capacity.
As can be seen in Section~\ref{sec:LSAM}, except for the insert locations in a neural network, the downsampling scheme that controls the complexity of the LightNL blocks is another important factor to be determined. We note that both insert locations and downsampling schedule of LightNL blocks are critical to the performance and computational cost of models. To automate the process of model design and find an optimal configuration of the proposed LightNL blocks, we propose an efficient Neural Architecture Search (NAS) method.
Concretely, 
we propose to jointly search the configurations of LightNL blocks and the basic neural network architectural parameters (\eg, kernel size, number of channels) using a cost-aware loss function.

\vspace{0.5ex}\noindent\textbf{Insert location.}~Motivated by~\cite{stamoulis2019single}, we select several candidate locations for inserting LightNL blocks throughout the network and decide whether a LightNL block should be used 
by
comparing the \textit{$L_2$ norm} of the depthwise convolution kernel $W_d$ to a trainable latent variable $t$:
\begin{equation} \label{eq:search_pos}
   \hat W_d = \mathbbm{1}\left(\|W_d\|^2 > t\right) \cdot W_d,
\end{equation}
where $\hat W_d$ replaces $W_d$ to be used in Eqn.~\eqref{eq:depthwise}, and $\mathbbm{1}(\cdot)$ is an indicator function.
$\mathbbm{1}(\cdot) = 1$ indicates that a LightNL block will be used  with $\hat W_d = W_d$ being the depthwise convolution kernel. Otherwise, $\hat W_d = 0$ when $\mathbbm{1}(\cdot) = 0$ and thus Eqn.~\eqref{eq:depthwise} is degenerated to $z=x$ meaning no lightweight non-local block will be inserted.

Instead of manually selecting the value of threshold $t$, we set it to be a trainable parameter, which is jointly optimized with other parameters via gradient decent. To compute the gradient of $t$, we relax the indicator function $I(x,t) = \mathbbm{1}(x>t)$ to a differentiable sigmoid function $\sigma(\cdot)$ during the back-propagation process.

\vspace{0.5ex}\noindent\textbf{Module compactness.}  As can be seen from Eqn.~\eqref{eq:remove_nl}, the computational cost of LightNL block when performing the matrix multiplication is determined by the compactness of downsampled features. 
Given a search space $R$ which contains $n$ candidate downsampling ratios,~\ie,~$R = \{r_1, r_2, r_3, ..., r_n\}$ where $0 \le r_1 < r_2 < r_3 < ... < r_n \le 1$, our goal is to search for an optimal downsampling ratio $r^*$ for each LightNL block. For the sake of clarity, here we use the case of searching downsampling ratios along the channel dimension to illustrate our method.
Note that searching downsampling ratios along other dimensions can be performed in the same manner.

Different from searching for the insert locations through Eqn.~\eqref{eq:search_pos}, we encode the choice of downsampling ratios in the process of computing affinity matrix:
\begin{equation} \label{eq:encode_spatial_downsample}
\begin{split}
    \ma{x_{att}} = & \mathbbm{1}(r_1) \cdot \ma{x_{r_1}}\ma{x_{r_1}^T} +  \mathbbm{1}(r_2) \cdot \ma{x_{r_2}}\ma{x_{r_2}^T} + ... \\
    & + \mathbbm{1}(r_{n-1}) \cdot \ma{x_{r_{n-1}}}\ma{x_{r_{n-1}}^T} + \mathbbm{1}(r_n) \cdot \ma{x_{r_n}}\ma{x_{r_n}^T},
\end{split}
\end{equation}
where $\ma{x_{att}}$ denotes the computed affinity matrix, $\ma{x}_{r}$ denotes the downsampled feature with downsampling ratio $r$, and $\mathbbm{1}(r)$ is an indicator which holds true when $r$ is selected. By setting the constraint that only  one downsampling ratio is used, Eqn.~\eqref{eq:encode_spatial_downsample} can be simplified as $\ma{x_{att}} = \ma{x_{r_i}}\ma{x_{r_i}^T}$ when $r_i$ is selected as the downsampling ratio.

A critical step is how to formulate the condition of $\mathbbm{1}(\cdot)$ for deciding which downsampling ratio to use. A reasonable intuition is that the criteria should be able to determine
whether the downsampled feature can be used to compute an accurate affinity matrix. Thus, our goal is to define a ``similarity" signal that models whether the affinity matrix from the downsampled feature is close to the ``ground-truth'' affinity matrix, denotes as $\ma{x_{gt}}\ma{x_{gt}^T}$. Specifically, we write the indicator as
\begin{equation} 
\label{eqn:search}
\begin{split}
    \mathbbm{1}(r_1) = & \mathbbm{1}\left(\|\ma{x_{r_1}}\ma{x_{r_1}^T} - \ma{x_{r_{gt}}}\ma{x_{gt}^T} \|^2 < t\right), \\
    \mathbbm{1}(r_2) = & \mathbbm{1}\left(\|\ma{x_{r_2}}\ma{x_{r_2}^T} - \ma{x_{gt}}\ma{x_{gt}^T} \|^2 < t\right) \land \neg \mathbbm{1}(r_1), \\
    & ...  \\
    \mathbbm{1}(r_n) = & \mathbbm{1}\left(\|\ma{x_{r_n}}\ma{x_{r_n}^T} - \ma{x_{gt}}\ma{x_{gt}^T} \|^2 < t\right) \\
    & \land \neg \mathbbm{1}(r_1) \land \neg \mathbbm{1}(r_2) \land ... \land \neg \mathbbm{1}(r_{n-1}),
\end{split}
\end{equation}
where $\land$ denotes the logical operator AND. An intuitive explanation to the rational of Eqn.~\eqref{eqn:search} is
the algorithm always selects the smallest $r$ with which the Euclidean distance between $ \ma{x_{r}}\ma{x_{r}}^T$ and $\ma{x_{gt}}\ma{x_{gt}}^T$ is lower than threshold $t$.
To ensure $\mathbbm{1}(r_n) = 1$ when all other indicators are zeros, we set $\ma{x_{gt}} = \ma{x_{r_n}}$ so that $\mathbbm{1}\left(\|\ma{x_{r_n}}\ma{x_{r_n}^T} - \ma{x_{gt}}\ma{x_{gt}^T} \|^2 < t\right) \equiv 1$. 
Meanwhile, we relax the indicator function to sigmoid when computing gradients and update the threshold $t$ via gradient descent. 
Since the output of indicator changes with different input feature $\ma{x}$, for better training convergence, we get inspired from batch normalization~\cite{ioffe2015batch} and use the exponential moving average of affinity matrices in computing Eqn.~\eqref{eqn:search}.
After the searching stage, the downsampling ratio is determined by evaluating the following indicators:
\begin{equation} 
\begin{split}
    \mathbbm{1}(r_1) = & \mathbbm{1}\left(\mathrm{EMA}(\|\ma{x_{r_1}}\ma{x_{r_1}^T} - \ma{x_{r_{gt}}}\ma{x_{gt}^T} \|^2) < t\right), \\
    & ...  \\
    \mathbbm{1}(r_n) = & \mathbbm{1}\left(\mathrm{EMA}(\|\ma{x_{r_n}}\ma{x_{r_n}^T} - \ma{x_{gt}}\ma{x_{gt}^T} \|^2) < t\right) \\
    & \land \neg \mathbbm{1}(r_1) \land \neg \mathbbm{1}(r_2) \land ... \land \neg \mathbbm{1}(r_{n-1}).
\end{split}
\end{equation}
where $\mathrm{EMA}(\ma{x})$ denotes the exponential moving averaged value of $\ma{x}$.

From Eqn.~(\ref{eqn:search}), one can observe that the 
output of indicator depends on indicators with smaller downsampling ratio. Based on this finding, we propose to reuse the affinity matrix computed with low-dimensional features (generated with lower downsampling ratios) when computing affinity matrix with high-dimensional features (generated with higher downsampling ratios).
Concretely, $\ma{x_{r_i}}$ can be partitioned into $[\ma{x_{r_{i-1}}}, \ma{x_{r_i\backslash r_{i-1}}}]$, $i > 1$. The calculation of affinity matrix using $\ma{x_{r_i}}$ can be decomposed as
\begin{equation}
\begin{split}
    \ma{x_{r_i}}\ma{x_{r_i}^T} = &
    \begin{bmatrix}
    \ma{x_{r_{i-1}}} & \ma{x_{r_i\backslash r_{i-1}}}
    \end{bmatrix}
    \begin{bmatrix}
    \ma{x_{r_{i-1}}^T} \\ \ma{x_{r_i\backslash r_{i-1}}^T}
    \end{bmatrix} \\
    = & \ \ma{x_{r_{i-1}}}\ma{x_{r_{i-1}}^T} \ + \  \ma{x_{r_i\backslash r_{i-1}}}\ma{x_{r_i\backslash r_{i-1}}^T},
\end{split}
\end{equation}
where $\ma{x_{r_{i-1}}}\ma{x_{r_{i-1}}^T}$ is the reusable affinity matrix computed with a smaller downsampling ratio (recall that $r_{i-1}<r_{i}$). This feature reusing paradigm can largely reduce the search overhead since computing affinity matrices with more choices of downsampling ratios does not incur any additional computation cost. The process of feature reusing is illustrated in Figure~\ref{fig:fig3}.

\vspace{0.5ex}\noindent\textbf{Searching process.}~We integrate our proposed search algorithm with Single-path NAS~\cite{stamoulis2019single} and jointly search basic architectural parameters (following MNasNet~\cite{tan2019mnasnet}) along with the insert locations and downsampling schemes of LightNL blocks.
We search downsampling ratios along both spatial and channel dimensions to achieve better compactness.
To learn efficient deep learning models, the overall objective function is to minimize both standard classification loss and the model's computation complexity which is related to both the insert locations and the compactness of LightNL blocks:
\begin{equation}\label{eq:loss}
   \min \mathcal{L}(\textbf{w}, \textbf{t}) = 
    CE( \textbf{w}, \textbf{t} ) + \lambda \cdot 
    \text{log}(CC(\textbf{t})),
\end{equation}
where $\textbf{w}$ denotes model weights and $\textbf{t}$ denotes architectural parameters which can be grouped in two categories: one is from LightNL block including the insert positions and downsampling ratios while the other follows MNasNet~\cite{tan2019mnasnet} including kernel size, number of channels,~\etc. $CE(\cdot)$ is the cross-entropy loss and $CC(\cdot)$ is the computation (\ie,~FLOPs) cost. We use gradient descent to optimize the above objective function in an end-to-end manner.

\section{Experiments} \label{sec:exp}
We first demonstrate the efficacy and efficiency of LightNL by manually inserting it into lightweight models in Section~\ref{sec:exp_lsam}. Then we apply the proposed search algorithm to the LightNL blocks in Section~\ref{sec:exp_search}. The evaluation and comparison with state-of-the-art methods are done on ImageNet classification~\cite{deng2009imagenet}.

\subsection{Manually Designed LightNL Networks}\label{sec:exp_lsam}
\noindent\textbf{Models.}~Our experiments are based on MobileNetV2 1.0~\cite{sandler2018mobilenetv2}. We insert LightNL blocks after the second $1\times1$ point-wise convolution layer in every MobileNetV2 block. We use $25\%$ channels to compute the affinity matrix for the sake of low computation cost. Also, if the feature map is larger than $14 \times 14$, we downsample it along the spatial axis with a stride of $2$. We call the transformed model MobileNetV2-LightNL for short. We compare the two models with different depth multipliers, including $0.5$, $0.75$, $1.0$ and $1.4$. 

\vspace{0.5ex}\noindent\textbf{Training setup.}
Following the training schedule in MNasNet~\cite{tan2019mnasnet}, we train the models using the synchronous training setup on $32$ Tesla-V100-SXM2-16GB GPUs. We use an initial learning rate of $0.016$, and a batch size of $4096$ (128 images per GPU). The learning rate linearly increases to $0.256$ in the first $5$ epochs and then is decayed by $0.97$ every $2.4$ epochs. We use a dropout of $0.2$, a weight decay of $1\mathrm{e-}5$ and Inception image preprocessing~\cite{szegedy2017inception} of size $224 \times 224$. Finally, we use exponential moving average on model weights with a momentum of $0.9999$. All batch normalization layers use a momentum of $0.99$.

\begin{figure}[tb]
\centering
\includegraphics[width=0.80\linewidth]{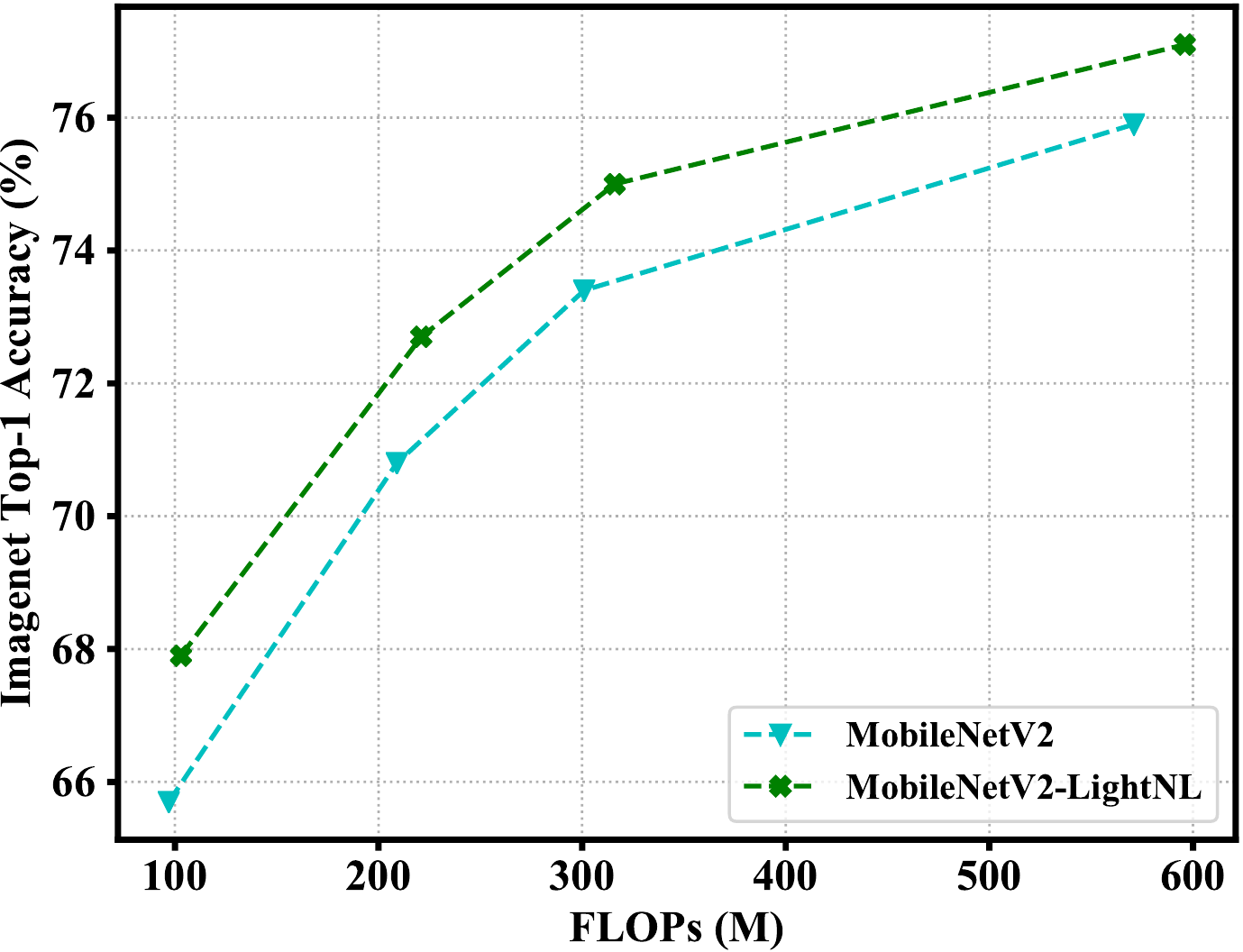}
\caption{\textbf{MobileNetV2 \emph{vs.} MobileNetV2-LightNL.}~We apply LightNL blocks to MobileNetV2 with different depth multipliers,~\ie,~$0.5$, $0.75$, $1.0$, $1.4$, from left to right respectively. Despite inserting LightNL blocks manually, consistent performance gains can be observed for different MobileNetV2 base models.}
\label{fig:mbv2}
\vspace{-1em}
\end{figure}

\vspace{0.5ex}\noindent\textbf{ImageNet classification results.}~We compare the results between the original MobileNetV2 and MobileNetV2-LightNL in Figure~\ref{fig:mbv2}. We observe consistent performance gain even without tuning the hyper-parameters of LightNL blocks for models with different depth multipliers. For example, when the depth multiplier is $1$, the original MobileNetV2 model achieves a top-1 accuracy of $73.4\%$ with $301$M FLOPs, while our MobileNetV2-LightNL achieves $75.0\%$ with $316$M FLOPs. According to Figure~\ref{fig:mbv2}, it is unlikely to boost the performance of the MobileNetV2 model to the comparable performance by simply increasing the width to get a $316$M FLOPs model.
When the depth multiplier is $0.5$, LightNL blocks bring a performance gain of $2.2\%$ with a marginal increase in FLOPs ($6$M).

\begin{figure*}[tb]
\centering
\includegraphics[width=0.8\linewidth]{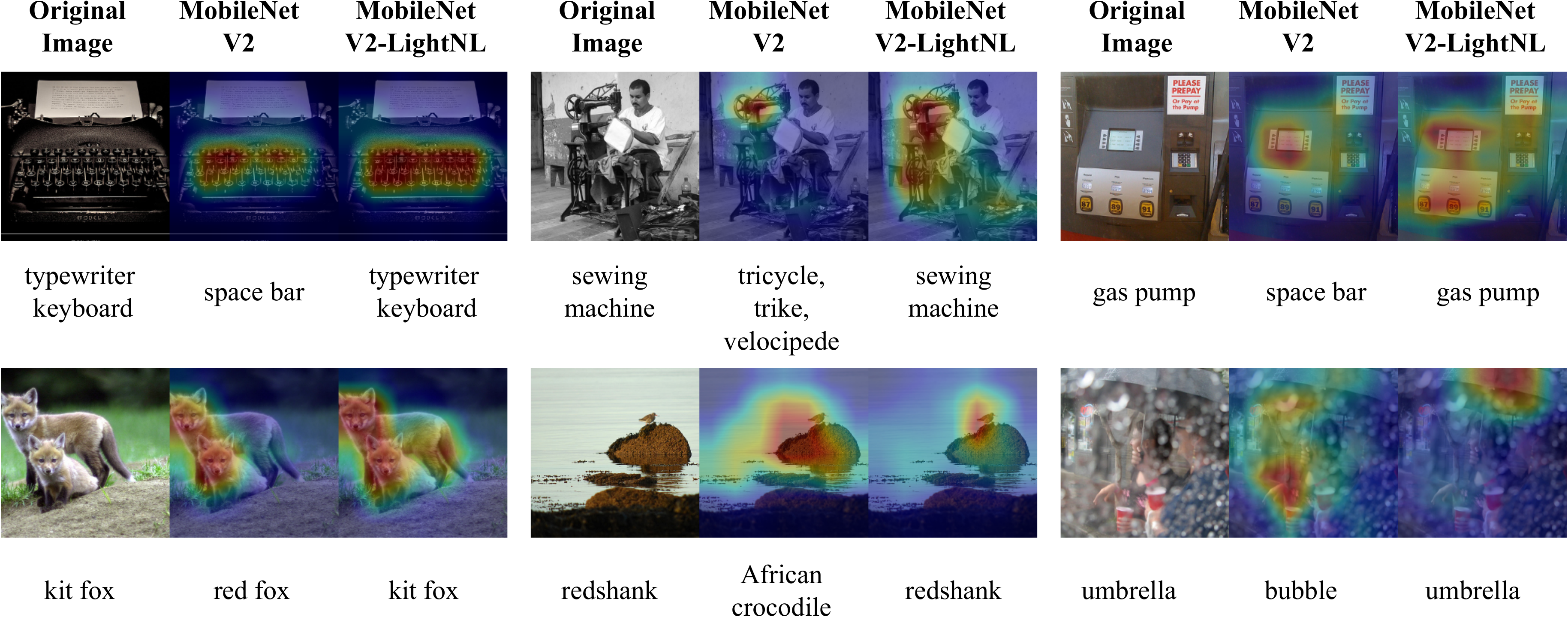}
\caption{Class Activation Map (CAM)~\cite{zhou2016learning} for MobileNetV2 and MobileNetV2-LightNL. The three columns correspond to the ground truth,  predictions by MobileNetV2 and predictions by MobileNetV2-LightNL respectively. The proposed LightNL block helps the model attend to image regions with more class-specific discriminative features.}
\label{fig:cam}
\vspace{-1em}
\end{figure*}

\begin{table}[tb]
\renewcommand\arraystretch{0.95}
\small
\centering
\begin{tabular}{llrr}
\toprule
\multicolumn{2}{c}{Non-local Module} & FLOPs / & \multicolumn{1}{l}{\multirow{2}{*}{Acc ($\%$)}} \\ \cline{1-2}
Operator & Wrapper & $\Delta$FLOPs & \multicolumn{1}{l}{} \\
\midrule
- & - & 301M & 73.4 \\ \hline
Wang~\etal~\cite{wang2018non} & \multirow{3}{*}{\begin{tabular}{@{}c@{}}Wang~\etal \\ \cite{wang2018non}\end{tabular}} & +6.2G & 75.2 \\
Levi~\etal~\cite{levi2018efficient} &  & +146M & 75.2 \\ 
Zhu~\etal~\cite{zhu2019asymmetric} &  & +107M & - \\ \hline
Eqn.~\eqref{eq:matrix_nl} &  \multirow{4}{*}{\begin{tabular}{@{}c@{}}Wang~\etal \\ \cite{wang2018non}\end{tabular}} & +119M & 75.2 \\
Eqn.~\eqref{eq:share_nl} &   & +93M & 75.1 \\
Eqn.~\eqref{eq:remove_nl} &   & +66M & 75.0 \\
Eqn.~\eqref{eq:compact_nl} &   & +38M & 75.0 \\ \hline
Eqn.~\eqref{eq:compact_nl} & Eqn.~\eqref{eq:depthwise}  & \textbf{+15M} & \textbf{75.0} \\
\bottomrule
\end{tabular}
\caption{\textbf{Ablation Analysis.} We present the comparison of different NL blocks and different variants in our design. The base model is MobileNetV2, which achieves a top-1 accuracy of $73.4$ with 301M FLOPs.}
\label{tab:ablation}
\vspace{-1em}
\end{table}

\begin{table}[tb]
\renewcommand\arraystretch{0.95}
\small
\centering
\begin{tabular}{lcc}
\toprule
Method   & FLOPs (M) & mIoU          \\
\midrule
MobileNetV2       & 301   & 70.6 \\
MobileNetV2-LightNL (\textbf{ours}) &   \textbf{316}  & \textbf{72.9} \\ 
\bottomrule
\end{tabular}
\caption{Comparison of FLOPs and mIoU on PASCAL VOC 2012.}
\label{tab:seg}
\vspace{-1em}
\end{table}

\vspace{0.5ex}\noindent\textbf{Ablation study.}~To diagnose the proposed LightNL block, we present a step-by-step ablation study in Table~\ref{tab:ablation}. As shown in the table, every modification preserves the model performance but reduces the computation cost. By comparing with the baseline model, the proposed LightNL block improves ImageNet top-1 accuracy by $1.6\%$ (from $73.4\%$ to $75.0\%$), but only increases $15$M FLOPs, which is only $5\%$ of the total FLOPs on MobileNetV2. Comparing with the standard NL block, the proposed LightNL block is about $400\times$ computationally cheaper (6.2G~\emph{vs.}~15M) with comparable performance ($75.2\%$~\emph{vs.}~$75.0\%$). Comparing with Levi~\etal~\cite{levi2018efficient} which optimized the matrix multiplication with the associative law, the proposed LightNL block is still $10\times$ computationally cheaper. Compared with a very recent work proposed by Zhu~\etal~\cite{zhu2019asymmetric} which leverages the pyramid pooling to reduce the complexity, LightNL is around $7\times$ computationally cheaper.

\vspace{0.5ex}\noindent\textbf{CAM visualization.} 
In order to illustrate the efficacy of our LightNL, Figure~\ref{fig:cam} compares the class activation map~\cite{zhou2016learning} for the original MobileNetV2 and MobileNetV2-LightNL. We see that LightNL is capable of helping the model to focus on more relevant regions while it is much computationally cheaper than the conventional counterparts as analyzed above. For example, at the middle top of Figure~\ref{fig:cam}, the model without the LightNL blocks focus on only a part of the sewing machine. When LightNL is applied, the model can ``see'' the whole machine, leading to more accurate and robust predictions.

\vspace{0.5ex}\noindent\textbf{PASCAL VOC segmentation results.}~To demonstrate the generalization ability of our method, we compare the performance of MobileNetV2 and MobileNetV2-LightNL on the PASCAL VOC 2012 semantic segmentation dataset~\cite{everingham2015pascal}.
Following Chen~\etal~\cite{chen2019renas}, we use the classification model as a drop-in replacement for the backbone feature extractor in the Deeplabv3~\cite{chen2017rethinking}. It is cascaded by an Atrous Spatial Pyramid Pooling module (ASPP)~\cite{chen2017deeplab} with three $3\times3$ convolutions with different atrous rates. The modified architectures share the same computation costs as the backbone models due to the low computation cost of LightNL blocks. All models are initialized with ImageNet pre-trained weights and then fine-tuned with the same training protocol in~\cite{chen2017deeplab}. 
It should be emphasized here that the focus of this part is to assess the efficacy of the proposed LightNL while keeping other factors fixed. It is notable that we do not adopt complex training techniques such as multi-scale and left-right flipped inputs, which may lead to better performance. The results are shown in Table~\ref{tab:seg}, LightNL blocks bring a performance gain of $2.3$ in mIoU with a minor increase in FLOPs.
The results indicate the proposed LightNL blocks are well suitable for other tasks such as semantic segmentation.

\begin{figure*}[tb]
    \centering
    \includegraphics[width=0.82\linewidth]{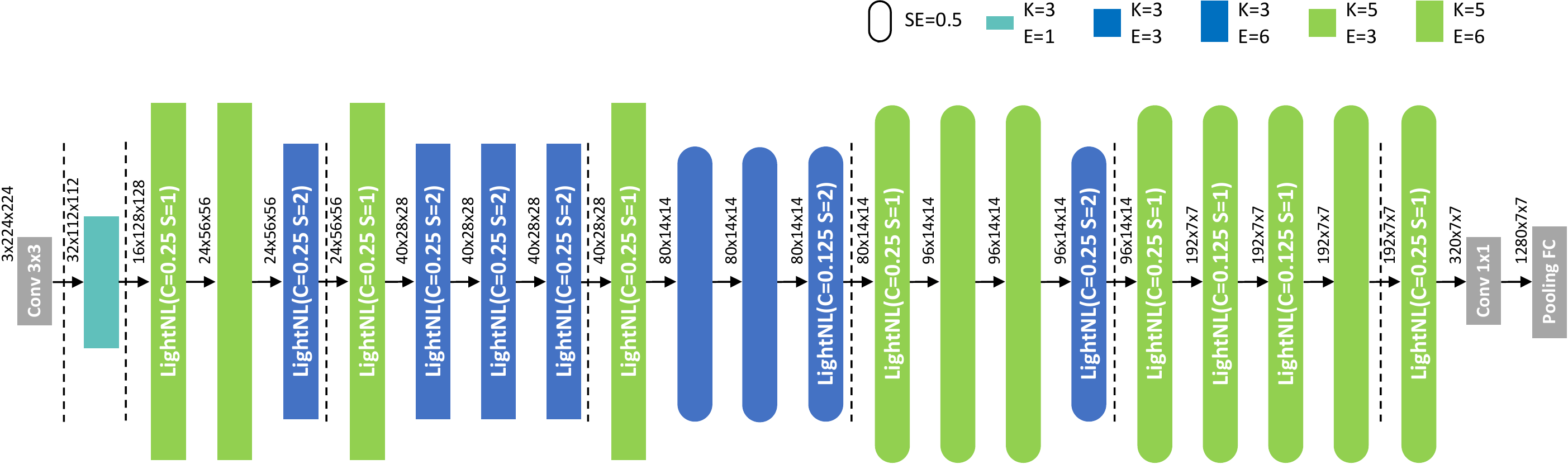}
    \caption{The searched architecture of AutoNL-L. C and S denote channel downsampling ratio and the stride of spatial downsampling respectively. We use different colors to denote the kernel size (K) of the depthwise convolution and use height to denote the expansion rate (E) of the block. We use the round corner to denote adding SE~\cite{hu2018se} to the MobileNetV2 block.}
    \label{fig:atomnas_m_arch}
    \vspace{-1.5em}
\end{figure*}

\begin{table}[tb]
\centering
\footnotesize
\begin{tabular}{lcccc}
\toprule
Model & \#Params & Flops & Top-1 & Top-5 \\
\midrule
MobileNetV2 \cite{sandler2018mobilenetv2}     & 3.4M & 300M & 72.0 & 91.0 \\
MBV2 (our impl.)       & 3.4M & 301M & 73.4 & 91.4 \\
ShuffleNetV2 \cite{ma2018shufflenet}         & 3.5M & 299M & 72.6 & - \\
\midrule
FBNet-A \cite{wu2019fbnet}          & 4.3M & 249M & 73.0 & - \\
Proxyless \cite{han2019proxyless}                   & 4.1M & 320M & 74.6 & 92.2 \\
MnasNet-A1~\cite{tan2019mnasnet}       & 3.9M & 312M & 75.2 & 92.5 \\
MnasNet-A2       & 4.8M & 340M & 75.6 & 92.7 \\
AA-MnasNet-A1~\cite{bello2019attention}       & 4.1M & 350M & 75.7 & 92.6 \\
MobileNetV3-L   & 5.4M & 217M & 75.2 & - \\
MixNet-S \cite{tan2019mixnet}       & 4.1M & 256M & 75.8 & 92.8 \\
\textbf{AutoNL-S (ours)}       & 4.4M & \textbf{267M} & \textbf{76.5} & \textbf{93.1} \\
\midrule
FBNet-C~\cite{wu2019fbnet}          & 5.5M & 375M & 74.9 & - \\
Proxyless (GPU)~\cite{han2019proxyless}                     & - & 465M & 75.1 & 92.5 \\
SinglePath~\cite{stamoulis2019single}  & 4.4M & 334M & 75.0 & 92.2 \\
SinglePath (our impl.)  & 4.4M & 334M & 74.7 & 92.2 \\
FairNAS-A~\cite{chu2019scarletnas}     & 4.6M & 388M & 75.3 & 92.4 \\
EfficientNet-B0~\cite{tan2019efficientnet}  & 5.3M & 388M & 76.3 & 93.2 \\
SCARLET-A \cite{chu2019scarletnas}  & 6.7M & 365M & 76.9 & 93.4 \\
MBV3-L (1.25x)~\cite{howard2019searching}    & 7.5M & 356M & 76.6 & - \\
MixNet-M~\cite{tan2019mixnet}       & 5.0M & 360M & 77.0 & 93.3 \\
\textbf{AutoNL-L (ours)}       & 5.6M & \textbf{353M} & \textbf{77.7} & \textbf{93.7} \\
\bottomrule
\end{tabular}
\caption{Comparison with the state-of-the-art models on ImageNet 2012 \texttt{Val} set.}\label{tab:overall_compare}
\vspace{-1em}
\end{table}

\subsection{AutoNL}\label{sec:exp_search}
We apply the proposed neural architecture search algorithm to search for an optimal configuration of LightNL blocks. Specifically, we have five LightNL candidates for each potential insert location,~\ie,~sampling $25\%$ or $12.5\%$ channels to compute affinity matrix, sampling along spatial dimensions with stride $1$ or $2$, inserting a LightNL block at the current position or not. Note that it is easy to enlarge the search space by including other LightNL blocks with more hyper-parameters. In addition, similar to recent work~\cite{tan2019mnasnet,wu2019fbnet,han2019proxyless,stamoulis2019single}, we also search for optimal kernel sizes, optimal expansion ratios and optimal SE ratios with MobileNetV2 block~\cite{sandler2018mobilenetv2} as the building block.

We directly search on the ImageNet training set and use a computation cost loss and the cross-entropy loss as guidance, both of which are differentiable thanks to the relaxations of the indicator functions during the back-propagation process. It takes $8$ epochs (about $32$ GPU hours) for the search process to converge. 

\vspace{0.5ex}\noindent\textbf{Performance on classification.}
We obtain two models using the proposed neural architecture search algorithm; we denote the large one as AutoNL-L and the small one as AutoNL-S in Table~\ref{tab:overall_compare}. The architecture of AutoNL-L is presented in Figure~\ref{fig:atomnas_m_arch}.

Table~\ref{tab:overall_compare} shows that AutoNL outperforms all the latest mobile CNNs. Comparing to the handcrafted models, AutoNL-S improves the top-1 accuracy by $4.5\%$ over MobileNetV2~\cite{sandler2018mobilenetv2} and $3.9\%$ over ShuffleNetV2~\cite{ma2018shufflenet} while saving about $10\%$ FLOPs. Besides, AutoNL achieves better results than the latest models from NAS approaches. For example, compared to EfficientNet-B0, AutoNL-L improves the top-1 accuracy by $1.4\%$ while saving about $10\%$ FLOPs. Our models also achieve better performance than the latest MobileNetV3~\cite{howard2019searching}, which is developed with several manual optimizations in addition to architecture search.

AutoNL-L also surpasses the state-of-the-art NL method (\ie,~AA-MnasNet-A1) by $2\%$ with comparable FLOPs. Even AutoNL-S improves accuracy by $0.8\%$ while saving $25\%$ FLOPs. We also compare with MixNet, which is a very recent state-of-the-art model under mobile settings, both AutoNL-L and AutoNL-S achieve $0.7\%$ improvement with comparable FLOPs but with much less search time ($32$ GPU hours~\emph{vs.}~$91,000$ GPU hours~\cite{wu2019fbnet}, $2,800\times$ faster).

\begin{figure}[t]
\centering
\includegraphics[width=0.7\linewidth]{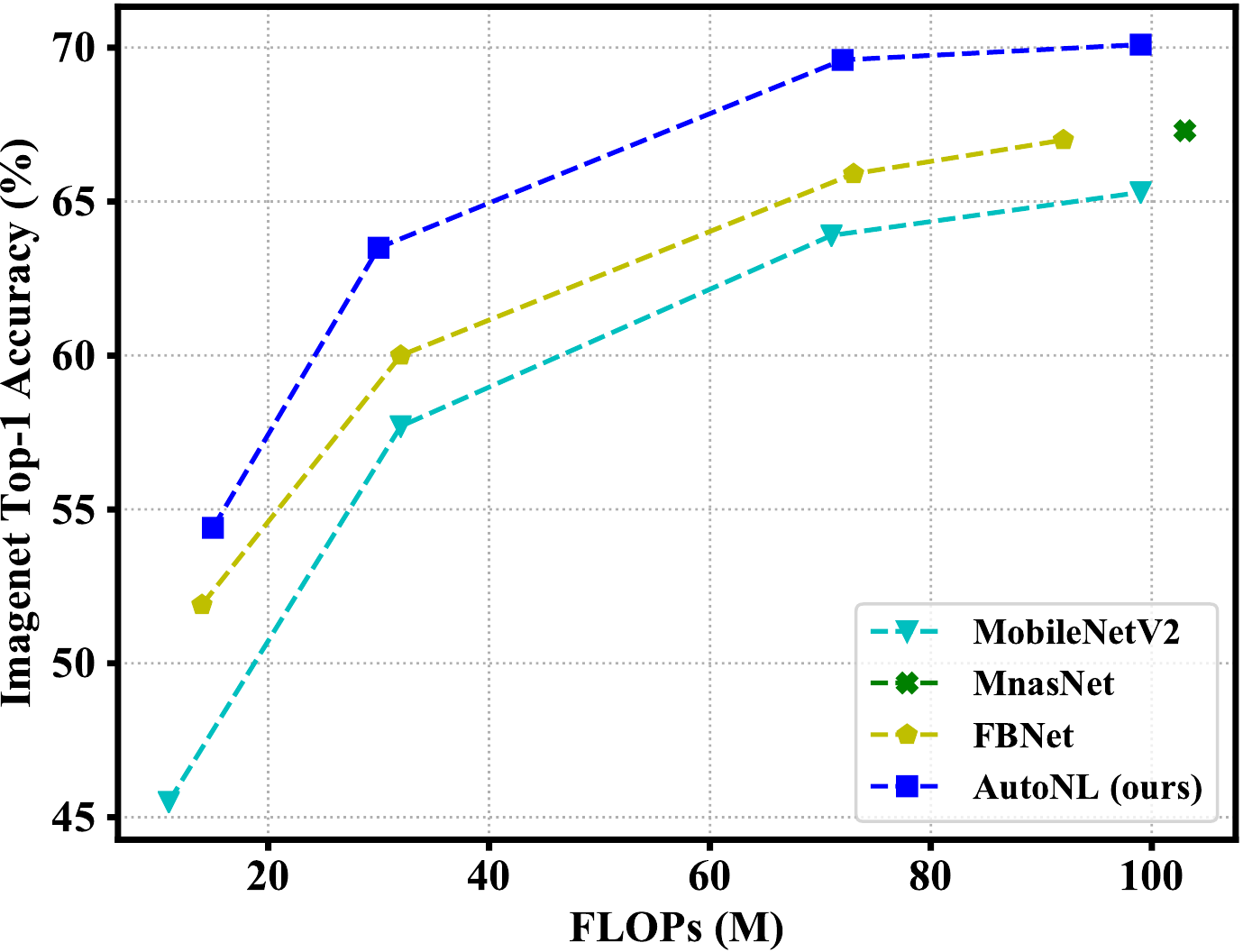}
\caption{Performance comparison on different input resolutions and depth multipliers under extremely low FLOPs. For MobileNetV2~\cite{sandler2018mobilenetv2}, FBNet~\cite{wu2019fbnet} and our searched models, the tuples of (input resolution, depth multiplier) are $(96, 0.35)$, $(128,0.5)$, $(192, 0.5)$ and $(128, 1.0)$ respectively from left to right. For MNasNet~\cite{tan2019mnasnet}, we show the result of $128$ input resolution with $1.0$ depth multiplier.} 
\label{fig:fbnet}
\vspace{-1.5em}
\end{figure}

We also search for models under different combinations of input resolutions and channel sizes under extremely low FLOPs. The results are summarized in Figure~\ref{fig:fbnet}. AutoNL achieves consistent improvement over MobileNetV2, FBNet, and MNasNet. For example, when the input resolution is $192$ and the depth multiplier is $0.5$, our model achieves $69.6\%$ accuracy, outperforming MobileNetV2 by $5.7\%$ and FBNet by $3.7\%$.

\section{Conclusion}
As an important building block for various vision applications, NL blocks under mobile settings remain underexplored due to their heavy computation overhead.
To our best knowledge, AutoNL is the first method to explore the usage of NL blocks for general mobile networks. 
Specifically, we design a LightNL block to enable highly efficient context modeling in mobile settings. We then propose a neural architecture search algorithm to optimize the configuration of LightNL blocks. Our method significantly outperforms prior arts with 77.7\% top-1 accuracy on ImageNet under a typical mobile setting (350M FLOPs).

\small{\vspace{1ex}\noindent\textbf{Acknowledgements}~This work was partially supported by ONR N00014-15-1-2356.}

{\small
\bibliographystyle{ieee_fullname}
\bibliography{egbib}
}

\end{document}